# Retinal Vessels Segmentation Based on Dilated Multi-Scale Convolutional Neural Network

**YUN JIANG[1], NING TAN[1], TINGTING PENG[1], AND HAI ZHANG[1].**
[1]College of Computer Science and Engineering, Northwest Normal University, Lanzhou, Gansu, China

This work was supported in part by the National Natural Science Foundation of China under Grant 61163036, in part by the program of NSFC Financing for Natural Science Fund in 2016 under Grant 1606RJZA047, in part by the Special Fund Project of Basic Scientific Research Operating Expenses of Institutes and Universities in Gansu in 2012, in part by Institutes and Universities Graduate Tutor Project in Gansu under Grant 1201-16, in part by the Third Period of the Key Scientific Research Project of Knowledge and Innovation Engineering of the Northwest Normal University under Grant nwnu-kjcxgc-03-67.

**ABSTRACT** Accurate segmentation of retinal vessels is a basic step in Diabetic retinopathy(DR) detection. Most methods based on deep convolutional neural network (DCNN) have small receptive fields, and hence they are unable to capture global context information of larger regions, with difficult to identify lesions. The final segmented retina vessels contain more noise with low classificaiton accuracy. Therefore, in this paper, we propose a DCNN structure named as D-Net. In the proposed D-Net, the dilation convolution is used in the backbone network to obtain a larger receptive field without losing spatial resolution, so as to reduce the loss of feature information and to reduce the difficulty of tiny thin vessels segmentation. The large receptive field can better distinguished between the lesion area and the blood vessel area. In the proposed Multi-Scale Information Fusion module (MSIF), parallel convolution layers with different dilation rates are used, so that the model can obtain more dense feature information and better capture retinal vessel information of different sizes. In the decoding module, the skip layer connection is used to propagate context information to higher resolution layers, so as to prevent low-level information from passing the entire network structure. Finally, our method was verified on DRIVE, STARE and CHASE dataset. The experimental results show that our network structure outperforms some state-of-art method, such as $N^4$-fields, U-Net, and DRIU in terms of accuracy, sensitivity, specificity, and $AUC_{ROC}$. Particularly, D-Net outperforms U-Net by 1.04 %, 1.23 % and 2.79 % in DRIVE, STARE, and CHASE three dataset, respectively.

## I. INTRODUCTION

RETINAL images have been widely used for diagnosis, screening and treatment of cardiovascular and ophthalmologic diseases [1], including two major diseases leading to blindness: age-related macular degeneration (AMD), diabetic retinopathy (DR) [2]. Vessel is a basic step required for the quantitative analysis of retinal images[3]. Due to the complex nature of retinal vessel network, the manual segmentation of vessels is a tedious task which also requires high skills. Automated retinal vessel segmentation has been widely studied over decades. however, it remains a challenging task.

The existing retinal vessel segmentation methods can be roughly divided into two main categories: unsupervised methods and supervised methods [1].

Unsupervised methods are designed according to the inherent characteristics of the blood vessels, do not require reference to manual annotations. In [3], a B-COSFIRE filter that selectively responds to blood vessels is proposed to automatically segment the vessel tree. In [4], use the zero-crossing characteristic inherent in Laplacian of Gaussian filter and combine the matched filter for retinal vessel extraction, which effectively avoids the mis-segmentation of the matched filter with Gaussian kernel. In [5], a matching filtering method based on the Gumbel probability distribution function is proposed to extract retinal blood vessels. In [6], proposed an automated method for retinal blood vessel segmentation using the combination of topological and morphological vessel extractors. In [7], used line de-tector filters and mathematical morphology was applied to extract retinal vessels. In [8], using a local entropy-based thresholding segmentation method for extract vascular tree structure. In [9], proposed a method extracting retinal blood vessels based on morphological component analysis (MCA) algorithm, which overcome producing false positive vessels. In [10], An unsupervised iterative blood vessel segmentation algorithm is employed to the fundus image. In [11], propose a trainable nonlinear filtering method called B-COSFIRE to segment the vessel tree. In [12], combines Gaussian smoothing, a morphological top-hat operator, and vessel contrast enhancement for background homogenization and noise reduction and then refined through curvature analysis and morphological reconstruction to segmentation retain vessel. In [13], Use some contrast-sensitive approaches to embedded traditional algorithms to improve the sensitivity of retinal vessel extraction. In [14], Propose a fully automatic filter method based on 3D rotating frames to segment retinal blood vessels.

Supervised methods can be further classified into two groups: 1). shallow learning based methods and 2). deep learning based methods. Generally, shallow learning based methods utilize handcrafted features for segmentation. Generally, shallow learning based methods utilize handcrafted features for segmentation. In [15], used based on the radial projection and semi-supervised method to extracting the retinal vessels. In [16], pixels are classified using a pixel neighborhood and a Gaussian mixture model (GMM) classifier. In [17], used based on a discriminatively trained fully connected conditional random field model for segmentation of the blood vessels in the fundus image. In [18], uses an ensemble system of bagged and boosted decision trees and utilizes a feature vector based on the orientation analysis of gradient vector field, morphological transformation, line strength measures, and Gabor filter responses to segmentation retina vessel. In contrast to shallow learning based methods, deep learning based methods automatically extract features for segmentation by training a large number of data samples. In [19], proposed a deep learning neural network (DNN) approach to segment the retinal vessels. In [20], the retinal vessel segmentation problem as a boundary detection task and solve it using deep learning and Conditional Random Field (CRF). In [21], using deep convolutional neural network training data augmentation samples achieved segmentation of blood vessels. In [22], the segmentation task as a multi-label task and utilize the implicit advantages of the combination of convolutional neural networks and structured prediction achieve the segmentation of blood vessels. In [23], a Size-Invariant Fully Convolutional Neural Network (SIFCN) is proposed to address the automatic retinal vessel segmentation problems. In [24], used a wide and deep neural network with strong induction ability to remolds the task of segmentation as a problem of cross-modality data transformation from the retinal image to vessel map. In [25], The blood vessels and optic discs are segmented using a deep convolutional neural network. In [26], proposed the skeletal similarity metric to be used as a pixelwise loss function for training deep learning models for retinal vessel segmentation. In [27], used a Recurrent Convolutional Neural Network (RCNN) based on U-Net as well as a Recurrent Residual Convolutional Neural Network (RRCNN) based on U-Net models to segmentation retina vessel. In [28], The convolutional neural network is sufficient utilized to extract high-level features and low-level features to segment retina vessels. In [29], used a connection sensitive attention U-Net(CSAU) for retinal vessel segmentation. In [30], a multi-level convolutional neural network supervised method is used to separate blood vessels from retina images and to distinguish small blood vessels by using local and global feature extractors.

Among these methods, traditional methods require prior knowledge and additional preprocessing to extra preprocess to extract hand-crafted feature information, and cannot obtain deeper feature information, which is susceptible to low-quality images and pathological regions. In the deep learning method that has been proposed, there are usually the following problems: 1). The downsampling factor of the model is too large, resulting in the loss of feature information of a large number of tiny thin vessels in the retinal image, which ultimately cannot be restored. 2). The receptive field of the model is too small, resulting in insufficient understanding of the local context information, and it is impossible to accurately distinguish the pathological regions and vessels in the retinal image, resulting in mis-segmentation. 3). The feature extraction ability of the network structure is insufficient, and it is difficult to restore low-level detail feature information, resulting in a lot of noises in the segmented blood vessel image. 4). It is not possible to accurately obtain vascular information of different sizes, resulting in inability to accurately detection the edges of vessels and tiny thin blood vessels.

In this paper, we proposed a retinal vessel segmentation model based on deep convolutional neural network. The main contributions of our work include:

1) We propose an automatic segmentation model for retinal vessels by D-Net, an end-to-end deep learning network. We use the residual module to improve the feature extraction ability of the network structure, reduce the downsampling factor to alleviate the excessive loss of feature information of tiny thin vessels, and using dilated convolution instead of the traditional convolution for dense sampling.

2) In the D-Net, we cascading the dilated convolution of different dilated rates to increase the receptive field of the kernel, which promotes the model to understanding of global context information, and effectively reduce the mis-segmentation of the pathological regions and retina vessels. Skip connection is used to promote the fusion of low-level detail information with high-level global context information, which alleviates the difficulty to restore the vessel edge and vessel feature information.

3) Multi-Scale Information Fusion module(MSIF) is pro-

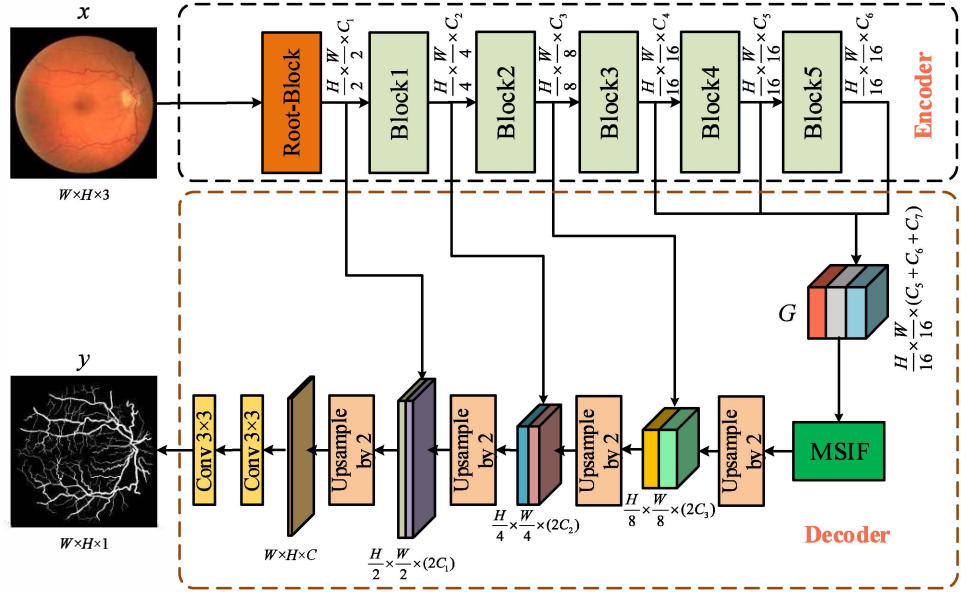

**FIGURE 1.** Retinal vessels image segmentation model.

posed, with the feature maps being sampled by using parallel convolution layers with different dilated rates, so as to obtain feature information of different scales, which improve the detection performance of the vessels edges and the tiny thin vessels.

## II. METHODS

In this section, we first expounded the theoretical knowledges of receptive field and dilated convolution [31], then introduce the network structure we proposed in detail.

### A. RECEPTIVE FIELD

We using different dilated ratios, the receptive field of the convolution filter can be changed. For a convolutional layer using a dilated convolution, if the dilated ratio is $r$, the convolution filter size is $k$. The receptive field size follows ((1)):

$$R_K = (k-1) \times (r-1) + k. \tag{1}$$

For instance, the convolutional layer uses a convolution kernel size of $3 \times 3$, a dilated rate $r = 4$, then the corresponding receptive field size 9.

Stacking multiple convolutional layers also allows for a larger receptive field. Suppose there are two convolutional layers, $L - 1$ and $L$, and the convolution filter are $k_1$ and $k_2$, respectively. The size of the receptive field of the $L + 1$ layer follows ((2)):

$$R_L = k_1 + k_2 - 1. \tag{2}$$

For instance, two stacked convolution layers, with their convolution filter size being 5 and 9 respectively, will result in a receptive field with its feature map being 13.

### B. DILATED CONVOLUTION

Compared to the traditional convolution operator. Dilated convolution is able to achieve a larger receptive field size without increasing the numbers of filter parameters and keeping the feature resolution unchanged. It is calculation formula is expressed as follows ((3)): $x[i]$ denotes the input signal, $y[i]$ denotes the output signal, $d$ is the dilation rate, $w[k]$ denotes the $k$-th parameter of filter, and $K$ is the filter size.

$$y[i] = \sum_{k=1}^{K} x[i + d \times k] \times w[k], \tag{3}$$

This equation reduces to a standard convolution when $d = 1$. Dilated convolution is equivalent to convolving of the input feature $x$ by inserting $d - 1$ zeros between two consecutive values of the convolution filter. For a convolution filter with size $k \times k$, the size of resulting dilated filter is $k_d \times k_d$, where $k_d = k + (k-1) \times (d-1)$. Thus, a large dilation rate has a large receptive field.

### C. D-NET ARCHITECTURE

D-Net is an end-to-end deep network model, which consists of three main parts. its network structure is shown in Fig. 1. The first part is the encoder, which is used to learn the feature information of the retinal image and rich hierarchical representation. The second part is Multi-Scale Information Fusion module (MSIF, please refer to Fig. 4), which capture multi-scale feature on top of the feature maps by using multiple parallel dilated convolutions with different dilated rates. The third part is the decoder, which is gradually upsampled by deconvolution on the feature map and finally restored to the same resolution as the input image $x$.

**TABLE 1.** Architectures for Network Backbone.

| Root-block | Block1 | | Block2 | | Block3 | | Block4 | | Block5 | |
|---|---|---|---|---|---|---|---|---|---|---|
| $3 \times 3 \times 32$ | $\begin{Bmatrix} 1 \times 1 \times 64 \\ 3 \times 3 \times 64 \\ 1 \times 1 \times 128 \end{Bmatrix}$ | $\times 3$ | $\begin{Bmatrix} 1 \times 1 \times 64 \\ 3 \times 3 \times 64 \\ 1 \times 1 \times 128 \end{Bmatrix}$ | $\times 3$ | $\begin{Bmatrix} 1 \times 1 \times 128 \\ 3 \times 3 \times 128 \\ 1 \times 1 \times 256 \end{Bmatrix}$ | $\times 3$ | $\begin{Bmatrix} 1 \times 1 \times 256 \\ 3 \times 3 \times 256 \\ 1 \times 1 \times 512 \end{Bmatrix}$ | $\times 3$ | $\begin{Bmatrix} 1 \times 1 \times 128 \\ 3 \times 3 \times 128 \\ 1 \times 1 \times 256 \end{Bmatrix}$ | $\times 3$ |
| $3 \times 3 \times 32$ | | | | | | | | | | |
| $3 \times 3 \times 64$ | | | | | | | | | | |
| $max\text{-}pool\ 3 \times 3$ | | | | | | | | | | |

## 1) Encoder

The structure of the encoder is shown in Table 1. Except for the first module, each of the other modules consists of three residual structures. Since the residual network [32] uses the short cut method, it becomes more sensitive to the change of weight, so that the network structure can make more fine adjustments to the weight. Each residual module consists of three convolution operations: $V \xrightarrow{1 \times 1} V^1 \xrightarrow{3 \times 3} V^2 \xrightarrow{1 \times 1} V^3$. For the input feature map $V$, the number of channels of $V$ is first reduced from $C^1$ to $C^2$ using a small convolution kernel ($1 \times 1$), and $V^1$ is output. Then use the large convolution kernel ($3 \times 3$) to extract the feature of $V^1$ and output $V^2$. Finally, use the small convolution kernel ($1 \times 1$) to restore the channel number from $C^2$ to $C^1$ and output $V^3$. Finally, the feature of $V$ and $V^3$ are added together to obtain the final result. The shortcut method adds the feature information of $V$ and $V^3$ to obtain the feature map $V'$, $V' = [V'_1, V'_2, ..., V'_c]$, where

$$V'_c = \sum_{i=1}^{W} \sum_{j=1}^{H} (V_c(i,j) + V_c^3(i,j)). \quad (4)$$

For convolution operations, we denote $F : X \rightarrow Y$, $X \in \mathbb{R}^{W^1 \times H^1 \times C^1}$, $Y \in \mathbb{R}^{W^2 \times H^2 \times C^2}$. For simplicity of exposition, in the notation that follows we take $F$ to be a standard convolutional operator. Let $K = [k_1, k_2, ..., k_{C^2}]$ denote the learned set of filter kernels, where $k_c$ refers to the parameters of the $c$-th filter. We can then write the outputs of $F$ as $Y = [y_1, y_2, ..., y_{C^2}]$, where

$$y_c = k_c * X = \sum_{n=1}^{C^1} k_c^n * x^n. \quad (5)$$

Here $*$ denotes convolution, $k_c = [k_c^1, k_c^2, ..., k_c^{C^1}]$, and $X = [x^1, x^2, ..., x^{C^1}]$.

In the retinal vessel segmentation, the color and brightness of some pathological regions are close to the vessels, which is easy to be mistakenly segmented. In order to reduce this mis-segmentation and make the model better understand the global context information, it is necessary to increase the receptive field of the model. However, if you use too much strided convolutions or pool to increase the receptive field, a lot of feature information will be lost, especially the tiny thin vessels in the retina image. So we reduce the downsampling factor, set it to 16, and then use the dilated convolution instead of the traditional convolution to increase the receptive field of the model. In the experiment, we replaced the standard convolution in block4 and block5 modules with dilated convolution, and each module consisted of 3 residual modules. We define the dilated rate of $Conv3 \times 3$ in these three residual modules as $(d_1, d_2, d_3)$, as shown in the Fig. 2. Finally, the feature maps outputted by block3, block4, and block5 are concatenated to obtain a feature map $G$, $G \in \mathbb{R}^{\frac{H}{16} \times \frac{W}{16} \times (C_5 + C_6 + C_7)}$

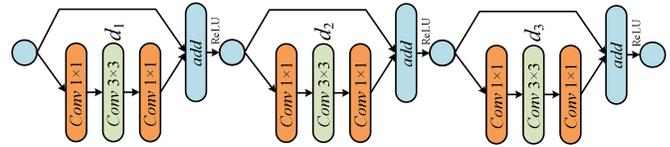

**FIGURE 2.** The detail structure of the Block.

Dilated convolution is constructed by inserting 'zeros' between each parameter in the convolutional kernel. For a convolution kernel with size $k \times k$, the size of resulting dilated filter is $k_d \times k_d$, where $k_d = k + (k-1) \times (d-1)$. For a dilated convolution with a convolution kernel size of $k_d \times k_d$, the effective value actually used for calculation is only $k \times k$. for instance, if $k = 3$, $d = 2$, $k_d = 5$, only 9 out of 25 pixels in the region are used for the computation. When the dilated rate $d$ is larger, the smaller proportion of the effective feature actually used for calculation which makes the more sparse the feature information captured by the model. In order to alleviate this problem, we reasonably set the values of different dilation rate $(d_1, d_2, d_3)$, where $d_1 < d_2 < d_3$. The dilated convolution of these different dilated rates is cascaded so that the final size of the receptive field fully covers a square region without any holes or missing edges. For simplicity of exposition, we use a one-dimensional graph to show, in which $(d_1, d_2, d_3) = (2, 2, 2)$ is set in Fig. 3 (a) and $(d_1, d_2, d_3) = (1, 2, 3)$ is set in Fig. 3 (b). It can be seen that in the two way, the receptive field obtained by the $L_4$ layer is the same, but in the Fig. 3 (a), the $L_4$ layer can only obtain a part of the feature information in the $L_1$ layer, and there are a large number of hole regions. In Fig. 3 (b), the $L_4$ layer obtains all the feature information in the $L_1$ layer.

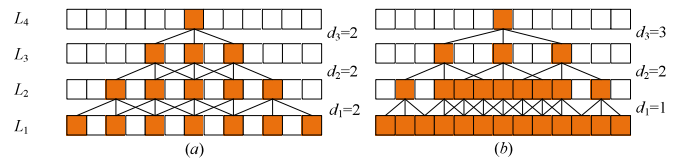

**FIGURE 3.** Sparse sampling and dense sampling of cascading dilated convolution.

## 2) Multi-scale information fusion module

The size of the blood vessels to be segmented in the retinal image is different. In order to better segment the retinal vessels of different sizes, we use the dilated convolution of different dilated rates in Multi-scale information fusion module (MSIF) for multi-scale feature capture to improve the detection accuracy of the vessels edges and the tiny thin vessels. The MSIF contains four parallel convolution layers and one global average pooling layer. The four parallel convolution layers are respectively three depthwise separable convolutions with different dilated rates and one $1 \times 1$ convolution layer. depthwise separable convolution [33], a powerful operation to reduce the computation cost and number of parameters while maintaining similar (or slightly better) performance. Where, three convolutional layers with different dilated rates can capture multi-scale context feature information, and the $1 \times 1$ convolution layer retains the feature information of the current scale. We use global average pooling to get image-level global context information [34] and then bilinearly upsample the feature to the desired spatial dimension, finally incorporate it into the model.

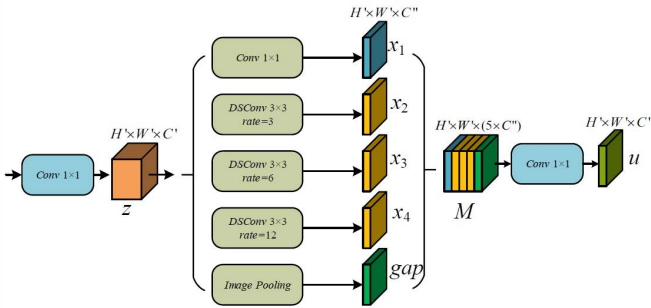

**FIGURE 4.** Multi-scale information fusion module detailed structure.

where the $c$-th element of image-level features gap is calculated by:

$$z_c = \frac{1}{W' \times H'} \sum_{i=1}^{W} \sum_{j=1}^{H} z_c(i, j) \qquad (6)$$

$$gap_c = F_{BI}(z_c) \qquad (7)$$

Here, $F_{BI}(\cdot)$ denotes bilinearly upsample, $z = [z_1, z_2, ..., z_{C'}]$ and $gap = [gap_1, gap_2, ..., gap_{C'}]$. Finally, the feature maps of all the branches are concatenated to obtain $M$, $M = [x_1, x_2, x_3, x_4, gap]$, and then uses a $[1 \times 1, 256]$ convolutional layer to fuse these multi-scale information to obtain the final feature map $u$.

## 3) Decoder

Decoder gradually upsampled the feature map $u$ output by the MSIF using deconvolution, with an upsampling factor of 2, and finally restores to the same resolution as the input image $x$. The Skip connection is used to concatenated the feature information after each deconvolution with the low-level detail information in the decoding layer, thereby alleviates that some thin-walled blood vessels and blood vessel edge information are difficult to recover during upsampling. Finally, the feature information is refined using two $3 \times 3$ convolution layers, and output the final segmentation result $y$.

## 4) Loss

In order to prevent over-fitting, we using the $L2$ regularization method to reduce over-fitting and improving the recognition ability of the convolutional layer. Dilated convolution is utilized in D-Net to expand the receptive field, and to take full advantage of context information for retinal vessel segmentation. The training of the whole network is formulated as a per-pixel classification problem with respect to the ground-truth segmentation masks, which is shown in ((8)):

$$\mathcal{L}(\chi; \theta) = \lambda \|W\|_2^2 - [\sum_{x \in \chi} \phi(x, \ell(x)) + \|y - \ell(\chi)\|], \quad (8)$$

Here, the first part is the regularization term, and the later one include target classifiers loss term and $L2$ distance in the training set. The tradeoff of these two terms is controlled by the hyperparameter $\lambda$. $W$ denotes the parameters for inferring the target output $y'$. Let $\phi(x, \ell(x))$ denotes the cross entropy loss regarding the true label $\ell(x)$ for pixel $x$ in image space $\chi$, $y$ denotes ground truth and $\ell(\chi)$ is the segmentation result predicted by the model. The parameters $\theta = \{W\}$ of deep contextual network are jointly optimized in an end-to-end way by minimizing the total loss function $\mathcal{L}(\chi; \theta)$.

# III. EXPERIMENT RESULTS

## A. DATASET

We validated our proposed method in three publicly available datasets, DRIVE, STARE and CHASE. The DRIVE dataset was obtained from a diabetic retinopathy screening program in The Netherlands. A total of 40 were selected from 400 subjects diabetic subjects aged between 25 and 90 years. Of these, 33 did not show any sign of diabetic retinopathy and 7 show signs. The training set and the test set each contains 20 sheets, and the size of each image is $565 \times 584$. (http://www.isi.uu.nl/Research/Databases/DRIVE/)

STARE database consists of 20 retinal fundus images, and each image was digitalized to $700 \times 605$ pixels. The first half of the dataset was collected by healthy subjects, while the other half pathological cases with abnormalities that overlap with blood vessels. In some cases obscuring them completely. The presence of lesions makes segmentation more challenging. STARE database contains two sets of manual segmentation prepared by two observers. (http://www.ces.clemson.edu/ ahoover/stare/)

CHASE contains 28 retinal images, collected from both the left and right eyes of 14 school children. With a resolution of $1280 \times 960$ pixels. Compared with DRIVE and STARE, images in CHASE have uneven background illumination, poor blood vessel contrast and extensive arteriolars. (https://blogs.kingston.ac.uk/retinal/chasedb1/)

### B. IMPLEMENTATION DETAILS

Our framework was implemented under the open-source deep learning library TensorFlow [35]. on a server with Intel(R) Xeon(R) E5-2620 v3 2.40GHz CPU, Tesla K80 GPU, and Ubuntu64 as OS. During training, the Adam optimizer [36] is used for gradient descent, with parameter setting: $\beta_1 = 0.9$, $\beta_2 = 0.999$, and $\varepsilon = 1e^{-8}$. The poly learning rate policy [31] is employed, with the initial learning rate being 0.0001. The learning rate during training is the initial learning rate multiplied by $(1 - \frac{iter}{max\_iter})^{power}$, with $power = 0.9$. The batch size is set to 4.

### C. PERFORMANCE EVALUATION

In order to evaluate the retinal vessels segmentation performance, we compared the performance by sensitivity, specificity, and accuracy, F1, which is widely used by the research community of image segmentation.

$$Accuracy = \frac{TP + TN}{TP + FN + TN + FP} \quad (9)$$

$$Precision = \frac{TP}{TP + FP} \quad (10)$$

$$Recall = \frac{TP}{TP + FN} \quad (11)$$

$$F1 = 2 \times Precison \times \frac{Recall}{Precison + Recall} \quad (12)$$

Here, $TP$ is the number of blood vessel pixels that are correctly segmented, $TN$ is the number of background pixels that are correctly segmented, $FP$ is the background pixel that is incorrectly segmented into blood vessel pixels, and $FN$ is a blood vessel pixel that is incorrectly marked as a background pixel.

### D. COMPARISON OF RESULTS BEFORE AND AFTER MODEL IMPROVEMENT

In Table 2, we compare the effects of setting different $(d_1, d_2, d_3)$ values on the performance of the model. AS show in the table, when setting $(d_1, d_2, d_3) = (1, 1, 1)$, it is the traditional convolution layer, the model's receptive field is the smallest, and the global context information cannot be fully understood. It is not possible to better distinguish between retinal vessel and pathological regions, resulting in the worst segmentation performance. When $(d_1, d_2, d_3) = (1, 2, 3)$ and $(d_1, d_2, d_3) = (1, 2, 4)$, the experimental results show that when $(d_1, d_2, d_3) = (1, 2, 4)$, the model's receptive field is the largest, the context information is more fully understood, and the model segmentation performance is also the best. Therefore, the receptive field of the model is increase by cascade the dilated convolution of different dilated rates, the segmentation performance of the model can be effectively improved.

In order to validate the effectiveness of the introducing the multi-scale information fusion module (MSIF), we compared the performance with and without MSIF module. It can be seen from the experimental results in Table 3 that the introduction of the MSIF module can make the network structure work better. Because MSIF can effectively capture multi-scale information so that the model can better segment retinal vessels of different size.

In Table 4, we compare the effect of setting different rates on the performance of the model for three parallel convolutional layers in the MSIF module. According to the result comparison, it can be seen that when the same $(d_1, d_2, d_3)$ value is set, the segmentation performance of the model is better when the range of local context information capture in the MSIF module is larger. For example, when $(d_1, d_2, d_3) = (1,2,4)$, the dilated rate in the MSIF module is set to $(3, 6, 12)$, the performance of the model is better than set to $(3, 5, 7)$.

### E. COMPARISON OF DIFFERENT METHODS

In this group of experiment, we compared the D-Net with some state-of-art methods, such as DRIU [25] and $N^4$-fields [37]. Fig. 5 and Fig. 6 compare the segmentation results of two retinal images from DRIVE and STARE dataset, respectively. In these figures, (a) presents the original retinal image, (b) presents ground true, (c) presents the segmentation result of DRIU, (d) presents the segmentation result of $N^4$-fields, and (e) presents the segmentation result by D-Net. In Fig. 5 and Fig. 6, the first row of images gives retinal image of a diabetic patient, and the second row gives that of a normal person. There is pathological region in the retinal image of a diabetic person, which is liable to cause mis-segmentation of the model. The retinal blood vessels segmented by the DRIU method contain a lot of noise, which forms many mis-segments, and the segmentation of small blood vessels is unclear and the boundary is blurred. Although the N4-fields method has less noise, it cannot distinguish the pathological from the retina vessel, and it is easy to mis-segmentation the pathological region into retina vessels, and the effect of segmentation the tiny thin retina vessels is not very good. In the method proposed in this paper, the dilated convolution is used instead of the pooling layer or the stride convolution, which reduces the loss of the feature information, thereby better recovering the tiny thin vessel information. Since the dilated convolution has a large receptive field, the pathological region can be well distinguished, so that the segmentation result is more accurate.

### F. QUANTITATIVE ANALYSIS OF DIFFERENT SEGMENTATION RESULTS

For further demonstrate the performance of D-Net for vessel segmentation, we evaluated D-Net with the previously proposed unsupervised method and supervised method on three datasets for sensitivity, Specificity, accuracy, $F1$ and other evaluation metric. Table 5, Table 6, and Table 7 show the segmentation results of different methods on the three data sets of DRIVE, STARE, and CHASE. As can be seen from the table, the supervised method generally better than the unsupervised method, and the deep learning method achieves particularly good results on the $AUC_{ROC}$.

**TABLE 2.** model with different $d_1, d_2, d_3$.

| $(d_1, d_2, d_3)$ | Receptive Field | F1 | Sensitivity | Specificity | Accuracy | $AUC_{ROC}$ | Time |
|---|---|---|---|---|---|---|---|
| (1, 1, 1) | 46 | 0.8021 | 0.7836 | 0.9753 | 0.9607 | 0.9714 | ∼1.0s |
| (1, 2, 3) | 116 | 0.8129 | **0.8047** | 0.9850 | 0.9671 | 0.9780 | ∼1.0s |
| (1, 2, 4) | 128 | **0.8132** | 0.7954 | **0.9849** | **0.9680** | **0.9800** | ∼1.0s |

**TABLE 3.** Effect of Multi-scale information fusion module.

| $(d_1, d_2, d_3)$ | use MSIF | Receptive Field | F1 | Sensitivity | Specificity | Accuracy | $AUC_{ROC}$ | Time |
|---|---|---|---|---|---|---|---|---|
| (1, 2, 4) | No | 128 | 0.8132 | 0.7954 | **0.9849** | 0.9680 | 0.9800 | **∼1.0s** |
| | Yes(3, 5, 7) | 142 | **0.8168** | **0.8181** | 0.9827 | **0.9681** | **0.9803** | ∼1.5s |

**TABLE 4.** Multi-scale information fusion module with cascade method.

| $(d_1, d_2, d_3)$ | MSIF | Receptive Field | F1 | Sensitivity | Specificity | Accuracy | $AUC_{ROC}$ | Time |
|---|---|---|---|---|---|---|---|---|
| | (3, 5, 7) | 131 | 0.8154 | 0.8171 | 0.9813 | 0.9654 | 0.9795 | ∼1.5s |
| (1, 2, 3) | (3, 6, 8) | 133 | 0.8149 | 0.8231 | 0.9809 | 0.9666 | 0.9805 | ∼1.5s |
| | (3, 6, 12) | 141 | 0.8172 | 0.8265 | 0.9816 | 0.9678 | 0.9819 | ∼1.5s |
| | (3, 5, 7) | 143 | 0.8168 | 0.8181 | 0.9827 | 0.9681 | 0.9803 | ∼1.5s |
| (1, 2, 4) | (3, 6, 8) | 145 | 0.8150 | **0.8275** | 0.9810 | 0.9673 | 0.9816 | ∼1.5s |
| | (3, 6, 12) | 153 | **0.8246** | 0.7839 | **0.9890** | **0.9709** | **0.9864** | ∼1.5s |

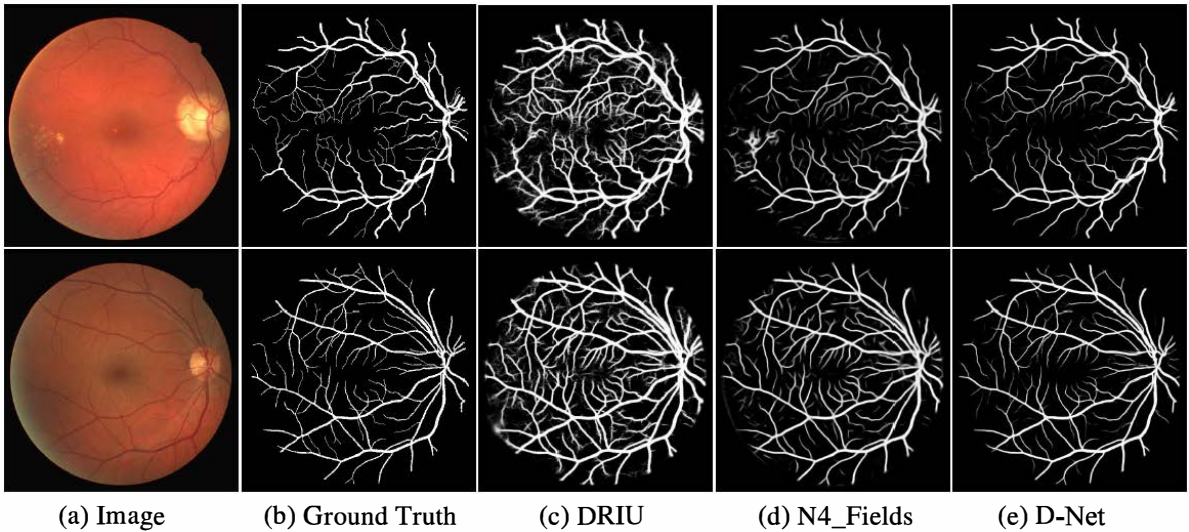

|  (a) Image | (b) Ground Truth | (c) DRIU | (d) N4_Fields | (e) D-Net |

**FIGURE 5.** Comparisons of segmentation results on DRIVE database. the first row is the retinal image of the diabetic patient, and the second row is the retinal image of the normal person.

On the DRIVE dataset, D-Net achieved a good result on all evaluation metric. The retina vessels segmented by the DRIU contain a lot of noise, and the segmented retina vessels are thicker than the actual retina vessels, and many background pixels are also segmented into retina vessels, so that the sensitivity is high and the specificity is low. The retinal vessels segmented by D-Net contain less noise and the segmentation results are more accurate. D-Net outperforms R2U-Net [27] by 0.75%, 0.8% and 0.78% in terms of $F1$,

$AUC_{ROC}$ and accuracy.

On the STARE data, the $F1$ of the D-Net method segmentation result is 84.92%, which is 0.17% higher than R2U-Net. The results of MSNN segmentation are higher than D-Net in sensitivity and $AUC_{ROC}$, but D-Net has the highest specificity and accuracy.

On the CHASE dataset, since the samples in the CHASE images have non-uniform background illumination, poor contrast of blood vessels and wider arteriolars that making the segmentation of retina vessels more difficult, and the model needs to have stronger feature extraction ability. However, D-Net outperforms other models in $AUC_{ROC}$, $F1$, accuracy and evaluation metric, of which $F1$ is 1.34% higher than R2U-Net. U-Net [38] has the highest sensitivity on this dataset, but not as good as D-Net in accuracy and $AUC_{ROC}$. Due to the large downsampling factor of R2U-Net, the tiny thin vessel feature information is seriously lost. The receptive field of R2U-Net is small, which makes it cannot obtain a large range of local feature information.

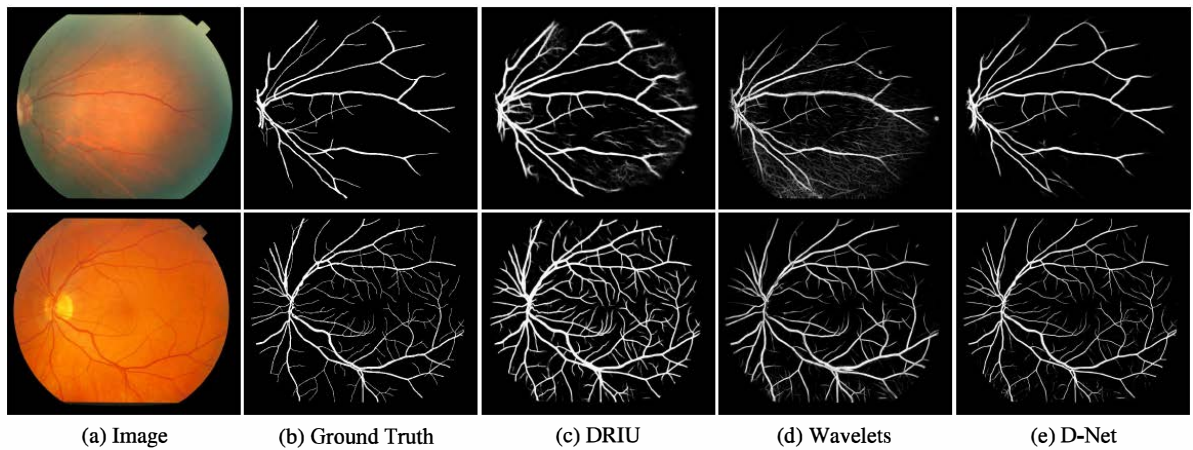

| (a) Image | (b) Ground Truth | (c) DRIU | (d) Wavelets | (e) D-Net |

**FIGURE 6.** Comparisons of segmentation results on STARE database. the first row is the retinal image of the diabetic patient, and the second row is the retinal image of the normal person.

Our proposed D-Net model uses a residual module in the backbone network to make finer adjustments to the weights so that the model can capture more useful feature information. We reduced the downsampling factor of the model which effectively alleviated the loss of tiny thin vessel feature information. Model cascaded the dilated convolution of different dilated rates for dense feature sampling, preventing excessive retina vessel feature information loss and maintaining the model's receptive field unchanged or larger. We propose that the MSIF module capture retina vessel feature information of different scales, which effectively improves the detection accuracy of the retina vessel edge information and tiny thin vessels. When upsampling, Skip connection is used to promote the recovery of detailed information of the retina vessel, which improves the accuracy of model segmentation. Through the analysis of the results on DRIVE, STARE, CHASE, which prove D-Net has better performance and robustness.

### G. EVALUATION OF ROC AND PR CURVES

In Fig. 7 and Fig. 8, we compare the Receiver Operating Characteristic (ROC) curve and Precision Recall (PR) curve of D-Net with several state-of-the-art retinal vessel segmentation method such as $N^4$-fields [37], Wavelet [42], DRIU [25], HED [43], and other methods. the ROC and PR curves area on the DRIU were 0.9861 and 0.8185, respectively. Although the $F1$ evaluation results of DRIU were not ideal in DRIVE, the area under the ROC and PR curves was comparable to our proposed D-Net. On the STARE dataset, the DRIU performed much better than the HED in the PR curve, however, it did not work as well as the HED in the $AUC_{ROC}$. The performance is unstable because the network structure used is simple and the feature extraction ability is relatively weak, which makes the generalization ability and robustness of the model relatively poor. Our D-Net obtains the best performances on the DRIVE dataset (0.9864 $AUC_{ROC}$) and the STARE dataset (0.9927 $AUC_{ROC}$), which has about 1% improvement on PR curve area than HED. It can be seen from the Fig. 7 and Fig. 8 that D-Net has better performance on the two data than other methods which prove D-Net has better feature extraction ability, generalization ability and robustness than other methods.

### IV. CONCLUSIONS

In this paper, we propose D-Net, an end-to-end deep convolutional neural network structure, for automatically segment retinal vessels. In the backbone network, samples of the downsampling layer is removed, which alleviates the problem that the loss of feature information is difficult to recover. The cascaded mode is emploied, which could increase the dilation rate of the convolution kernel gradually, so as to keep the receptive field unchanged (or larger). In the MSIF module, the parallel dilation convolution of different dilation ratios is used to perform dense information sampling on the feature map, and the retinal vessel information of different sizes is better captured. In order to reduce the parameters and increase the speed of the model, the depthwise separable convolution is used instead of the standard convolution in MSIF. Due to some low-level information is difficult to recover, skip layer connection is utilized to directly fusion low-level information and high-level information in the network structure. Finally, our method was verified on DRIVE, STARE and CHASE dataset, and the experiment result show that the proposed algorithm has better performance for retinal vessel segmentation than some state-of-art algorithms, such as N4-fields, U-Net, and DRIU.

### ACKNOWLEDGMENT

The authors would like to thank anonymous reviewers for their constructive suggestions.

**TABLE 5.** Comparison of proposed methods with other methods in the DRIVE database.

| Type | Methods | Year | F1 | Sensitivity | Specificity | Accuracy | AUC$_{ROC}$ | Time |
|------|---------|------|-----|-------------|-------------|----------|-------------|------|
| Unsupervised methods | Lam [39] | 2010 | - | - | - | 0.9472 | 0.9614 | ~13m |
| | Fraz [1] | 2011 | - | 0.7152 | 0.9759 | 0.9430 | - | ~2m |
| | You [40] | 2011 | - | 0.7410 | 0.9751 | 0.9434 | - | - |
| | Azzopardi [3] | 2015 | - | 0.7655 | 0.9704 | 0.9442 | 0.9614 | ~10s |
| Supervised methods | Marin [41] | 2011 | - | 0.7067 | 0.9801 | 0.9452 | 0.9558 | ~90s |
| | Fraz [1] | 2012 | - | 0.7406 | 0.9807 | 0.9480 | 0.9747 | ~100s |
| | Roychowdhury [16] | 2016 | - | 0.7250 | 0.9830 | 0.9520 | 0.9620 | ~6.5s |
| | Liskowsk [21] | 2016 | - | 0.7763 | 0.9768 | 0.9495 | 0.9720 | - |
| | Qiaoliang Li [24] | 2016 | - | 0.7569 | 0.9816 | 0.9527 | 0.9738 | ~4.0s |
| | DRIU [25] | 2016 | 0.8210 | **0.8261** | 0.9115 | 0.9541 | 0.9861 | ~3.0s |
| | U-Net [38] | 2018 | 0.8142 | 0.7537 | 0.9820 | 0.9531 | 0.9755 | ~4.0s |
| | R2U-Net [27] | 2018 | 0.8171 | 0.7792 | 0.9813 | 0.9556 | 0.9784 | ~5.0s |
| | MSNN [28] | 2018 | - | 0.8033 | 0.9808 | 0.9581 | 0.9826 | ~3.0s |
| | D-Net(Ours) | 2019 | **0.8246** | 0.7839 | **0.9890** | **0.9709** | **0.9864** | **~1.5s** |

**TABLE 6.** Comparison of proposed methods with other methods in the STARE database.

| Type | Methods | Year | F1 | Sensitivity | Specificity | Accuracy | AUC$_{ROC}$ | Time |
|------|---------|------|-----|-------------|-------------|----------|-------------|------|
| Unsupervised methods | Lam [39] | 2010 | - | - | - | 0.9567 | 0.9739 | ~13m |
| | Fraz [1] | 2011 | - | 0.7311 | 0.9680 | 0.9442 | - | ~100s |
| | You [40] | 2011 | - | 0.7260 | 0.9756 | 0.9497 | - | - |
| | Azzopardi [3] | 2015 | - | 0.7716 | 0.9701 | 0.9563 | 0.9497 | ~11s |
| Supervised methods | Marin [41] | 2011 | - | 0.6940 | 0.9770 | 0.9520 | 0.9820 | ~90s |
| | Fraz [1] | 2012 | - | 0.7548 | 0.9763 | 0.9534 | 0.9768 | ~100s |
| | Roychowdhury [16] | 2016 | - | 0.7720 | 0.9730 | 0.9510 | 0.9690 | ~14s |
| | Liskowsk [21] | 2016 | - | 0.7867 | 0.9754 | 0.9566 | 0.9785 | - |
| | Qiaoliang Li [24] | 2016 | - | 0.7726 | 0.9844 | 0.9628 | 0.9879 | ~4.5s |
| | DRIU [25] | 2016 | 0.7385 | 0.6066 | **0.9956** | 0.9499 | 0.9896 | ~6.5s |
| | U-Net [38] | 2018 | 0.8373 | 0.8270 | 0.9842 | 0.9690 | 0.9898 | ~7.8s |
| | R2U-Net [27] | 2018 | 0.8475 | 0.8298 | 0.9862 | 0.9712 | 0.9914 | ~7.5s |
| | MSNN [28] | 2018 | - | **0.8579** | 0.9826 | 0.9732 | **0.9930** | ~4.0s |
| | CSAU [29] | 2019 | 0.8435 | 0.8465 | - | 0.9673 | 0.9834 | - |
| | D-Net(Ours) | 2019 | **0.8492** | 0.8249 | 0.9904 | **0.9781** | 0.9927 | **~2.0s** |

**TABLE 7.** Comparison of proposed methods with other methods in the CHASE database.

| Type | Methods | Year | F1 | Sensitivity | Specificity | Accuracy | AUC$_{ROC}$ | Time |
|------|---------|------|-----|-------------|-------------|----------|-------------|------|
| Unsupervised methods | Azzopardi [3] | 2015 | - | 0.7716 | 0.9701 | 0.9563 | 0.9497 | - |
| Supervised methods | Fraz [1] | 2012 | - | 0.7224 | 0.9711 | 0.9469 | 0.9712 | ~120s |
| | Roychowdhury [16] | 2016 | - | 0.7201 | 0.9824 | 0.9530 | 0.9532 | ~12s |
| | Qiaoliang Li [24] | 2016 | - | 0.7507 | 0.9793 | 0.9581 | 0.9793 | - |
| | U-Net [38] | 2018 | 0.7783 | **0.8288** | 0.9701 | 0.9578 | 0.9772 | ~8.1s |
| | R2U-Net [27] | 2018 | 0.7928 | 0.7756 | 0.9820 | 0.9634 | 0.9815 | ~7.5s |
| | MSNN [28] | 2018 | - | 0.7742 | 0.9876 | 0.9662 | 0.9865 | ~4.4s |
| | D-Net(Ours) | 2018 | **0.8062** | 0.7839 | **0.9894** | **0.9721** | **0.9866** | **~2.1s** |

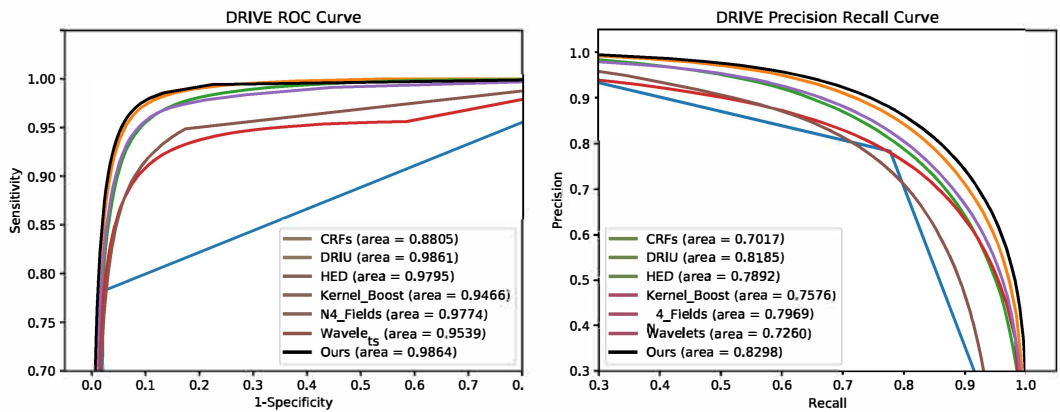

**FIGURE 7.** Receiver Operating Characteristic (ROC) curve and Precision Recall (PR) curve for various methods on DRIVE dataset.

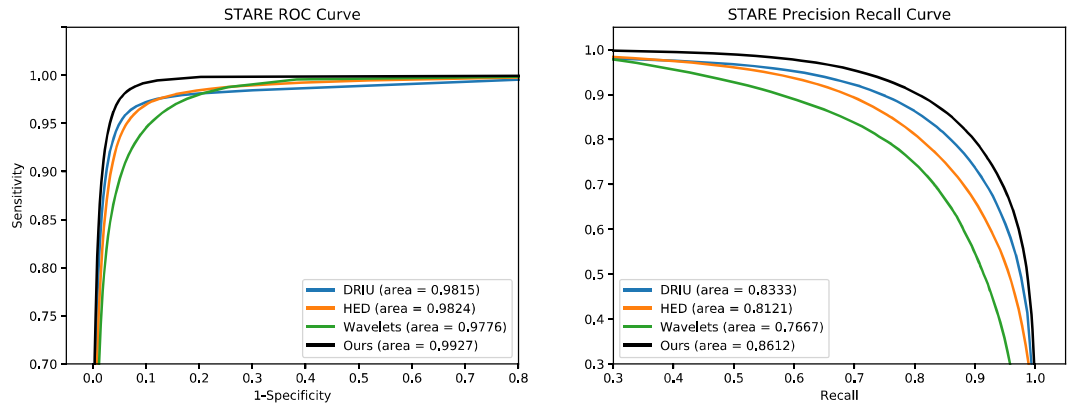

**FIGURE 8.** Receiver Operating Characteristic (ROC) curve and Precision Recall (PR) curve for various methods on STARE dataset.

## REFERENCES

[1] M. M. Fraz et al., "Blood vessel segmentation methodologies in retinal images–a survey," Computer Methods & Programs in Biomedicine, vol. 108, no. 1, pp. 407-433, 2012.

[2] M. D. Abrld'moff et al., "Automated analysis of retinal images for detection of referable diabetic retinopathy," Jama Ophthalmology, vol. 131, no. 3, p. 351, 2013.

[3] G. Azzopardi, N. Strisciuglio, M. Vento, and N. Petkov, "Trainable COS-FIRE filters for vessel delineation with application to retinal images," Medical Image Analysis, vol. 19, no. 1, pp. 46-57, 2015.

[4] D. Kumar, A. Pramanik, S. S. Kar, and S. P. Maity, "Retinal blood vessel segmentation using matched filter and Laplacian of Gaussian," in 2016 International Conference on Signal Processing and Communications (SPCOM), 2016, pp. 1-5.

[5] N. P. Singh and R. Srivastava, "Retinal blood vessels segmentation by using Gumbel probability distribution function based matched filter." Computer Methods & Programs in Biomedicine, vol. 129, no. C, pp. 40-50, 2016.

[6] J. Rodrigues and N. Bezerra, "Retinal Vessel Segmentation Using Parallel Grayscale Skeletonization Algorithm and Mathematical Morphology," in 2016 29th SIBGRAPI Conference on Graphics, Patterns and Images (SIBGRAPI), 2016, pp. 17-24.

[7] R. Aramesh and K. Faez, "A new method for segmentation of retinal blood vessels using Morphological image processing technique," International Journal of Advanced Studies in Computer Science and Engineering, vol. 3, no. 1, 2014.

[8] G. Hamednejad and H. Pourghassem, "Retinal blood vessel classification based on color and directional features in fundus image," in 2015 22nd Iranian Conference on Biomedical Engineering (ICBME), 2015, pp. 257-262.

[9] E. Imani, M. Javidi, and H. R. Pourreza, "Improvement of retinal blood vessel detection using morphological component analysis," Comput Methods Programs Biomed, vol. 118, no. 3, pp. 263-279, 2015.

[10] S. Roychowdhury, D. D. Koozekanani, and K. K. Parhi, "Iterative Vessel Segmentation of Fundus Images," IEEE Transactions on Biomedical Engineering, vol. 62, no. 7, pp. 1738-1749, 2015.

[11] N. Strisciuglio, M. Vento, G. Azzopardi, and N. Petkov, "Unsupervised delineation of the vessel tree in retinal fundus images," Computational Vision and Medical Image Processing: VIPIMAGE, vol. 1, pp. 149-155, 2015.

[12] L. C. Neto, G. L. B. Ramalho, J. F. S. R. Neto, R. M. S. Veras, and F. N. S. Medeiros, "An unsupervised coarse-to-fine algorithm for blood vessel segmentation in fundus images," Expert Systems with Applications, vol. 78, no. C, pp. 182-192, 2017.

[13] Y. Zhao, L. Rada, K. Chen, S. P. Harding, and Y. Zheng, "Automated Vessel Segmentation Using Infinite Perimeter Active Contour Model with Hybrid Region Information with Application to Retinal Images," IEEE Transactions on Medical Imaging, vol. 34, no. 9, pp. 1797-1807, 2015.

[14] J. Zhang, B. Dashtbozorg, E. Bekkers, J. P. W. Pluim, R. Duits, and B. M. t. H. Romeny, "Robust Retinal Vessel Segmentation via Locally Adaptive Derivative Frames in Orientation Scores," IEEE Transactions on Medical Imaging, vol. 35, no. 12, pp. 2631-2644, 2016.

[15] X. You, Q. Peng, Y. Yuan, CHEUNG, YiuMing, and J. Lei, "Segmentation of retinal blood vessels using the radial projection and semi-supervised approach," Pattern Recognition, vol. 44, no. 10, pp. 2314-2324, 2011.

[16] S. Roychowdhury, D. D. Koozekanani, and K. K. Parhi, "Blood Vessel Segmentation of Fundus Images by Major Vessel Extraction and Subimage Classification," IEEE Journal of Biomedical & Health Informatics, vol. 19, no. 3, pp. 1118-1128, 2017.

[17] J. Orlando, E. Prokofyeva, and M. Blaschko, "A Discriminatively Trained Fully Connected Conditional Random Field Model for Blood Vessel Segmentation in Fundus Images," IEEE transactions on bio-medical engineering, vol. 64, no. 1, pp. 16-27, 2016.

[18] M. M. Fraz et al., "An Ensemble Classification-Based Approach Applied to Retinal Blood Vessel Segmentation," IEEE Transactions on Biomedical Engineering, vol. 59, no. 9, pp. 2538-2548, 2012.

[19] M. Melinscak, P. Prentasic, and S. Loncaric, "Retinal vessel segmentation using deep neural networks," In Proceedings of 10th International Conference on Computer Vision Theory and Applications (VISAPP 2015), 2015, pp. 577-582.

[20] H. Fu, Y. Xu, S. Lin, D. Wing Kee Wong, and J. Liu, DeepVessel: Retinal Vessel Segmentation via Deep Learning and Conditional Random Field. 2016, pp. 132-139.

[21] P. Liskowski and K. Krawiec, "Segmenting Retinal Blood Vessels With Deep Neural Networks," IEEE Transactions on Medical Imaging, vol. 35, no. 11, pp. 2369-2380, 2016.

[22] A. Dasgupta and S. Singh, "A fully convolutional neural network based structured prediction approach towards the retinal vessel segmentation," in 2017 IEEE 14th International Symposium on Biomedical Imaging (ISBI 2017), 2017, pp. 248-251.

[23] Y. Luo, H. Cheng, and L. Yang, "Size-Invariant Fully Convolutional Neural Network for vessel segmentation of digital retinal images," in 2016 Asia-Pacific Signal and Information Processing Association Annual Summit and Conference (APSIPA), 2016, pp. 1-7.

[24] Q. Li, B. Feng, L. Xie, P. Liang, H. Zhang, and T. Wang, "A Cross-Modality Learning Approach for Vessel Segmentation in Retinal Images," IEEE Transactions on Medical Imaging, vol. 35, no. 1, pp. 109-118, 2015.

[25] K. K. Maninis, J. Pont-Tuset, P. Arbelécez, and L. V. Gool, "Deep Retinal Image Understanding," in International Conference on Medical Image Computing and Computer-Assisted Intervention, 2016, pp. 140-148.

[26] Z. Yan, X. Yang, and K. T. Cheng, "A Skeletal Similarity Metric for Quality Evaluation of Retinal Vessel Segmentation," IEEE Transactions on Medical Imaging, vol. 37, no. 4, p. 1045, 2018.

[27] M. Z. Alom, M. Hasan, C. Yakopcic, T. M. Taha, and V. K. Asari, "Recurrent Residual Convolutional Neural Network based on U-Net (R2U-Net) for Medical Image Segmentation," 2018.

[28] B. Zhang, S. Huang, and S. Hu, "Multi-scale Neural Networks for Retinal Blood Vessels Segmentation," 2018.

[29] R. Li, M. Li, and J. Li, "Connection Sensitive Attention U-NET for

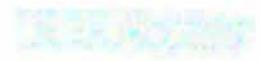

Accurate Retinal Vessel Segmentation," arXiv preprint arXiv:1903.05558, 2019.

[30] J. Guo, S. Ren, Y. Shi, and H. Wang, "Automatic Retinal Blood Vessel Segmentation Based on Multi-Level Convolutional Neural Network," in 2018 11th International Congress on Image and Signal Processing, BioMedical Engineering and Informatics (CISP-BMEI), 2018, pp. 1-5.

[31] L.C. Chen, G. Papandreou, I. Kokkinos, K. Murphy, and A. L. Yuille, "Deeplab: Semantic image segmentation with deep convolutional nets, atrous convolution, and fully connected crfs," IEEE transactions on pattern analysis and machine intelligence, vol. 40, no. 4, pp. 834-848, 2018.

[32] K. He, X. Zhang, S. Ren, and J. Sun, "Deep residual learning for image recognition," in Proceedings of the IEEE conference on computer vision and pattern recognition, 2016, pp. 770-778.

[33] A. G. Howard et al., "Mobilenets: Efficient convolutional neural networks for mobile vision applications," arXiv preprint arXiv:1704.04861, 2017.

[34] H. Zhao, J. Shi, X. Qi, X. Wang, and J. Jia, "Pyramid scene parsing network," in Proceedings of the IEEE conference on computer vision and pattern recognition, 2017, pp. 2881-2890.

[35] M. Abadi, A. Agarwal, P. Barham, E. Brevdo, and X. Zheng, "TensorFlow: Large-Scale Machine Learning on Heterogeneous Distributed Systems," 2016.

[36] D. P. Kingma and J. Ba, "Adam: A method for stochastic optimization," arXiv preprint arXiv:1412.6980, 2014.

[37] Y. Ganin and V. Lempitsky, "$N^4$-Fields: Neural Network Nearest Neighbor Fields for Image Transforms," in Asian Conference on Computer Vision, 2014, pp. 536-551: Springer.

[38] O. Ronneberger, P. Fischer, and T. Brox, "U-Net: Convolutional Networks for Biomedical Image Segmentation," in International Conference on Medical Image Computing & Computer-assisted Intervention, 2015.

[39] B. S. Y. Lam, Y. Gao, and A. W. Liew, "General Retinal Vessel Segmentation Using Regularization-Based Multiconcavity Modeling," IEEE Transactions on Medical Imaging, vol. 29, no. 7, pp. 1369-1381, 2010.

[40] X. You, Q. Peng, Y. Yuan, Y.m. Cheung, and J. Lei, "Segmentation of retinal blood vessels using the radial projection and semi-supervised approach," Pattern Recognition, vol. 44, no. 10-11, pp. 2314-2324, 2011.

[41] D. Marin, A. Aquino, M. E. Gegundez-Arias, and J. M. Bravo, "A New Supervised Method for Blood Vessel Segmentation in Retinal Images by Using Gray-Level and Moment Invariants-Based Features," IEEE Transactions on Medical Imaging, vol. 30, no. 1, pp. 146-158, 2011.

[42] S. Dua, U. R. Acharya, P. Chowriappa, and S. V. Sree, "Wavelet-Based Energy Features for Glaucomatous Image Classification," IEEE Transactions on Information Technology in Biomedicine, vol. 16, no. 1, pp. 80-87, 2012.

[43] S. Xie and Z. Tu, "Holistically-Nested Edge Detection," in 2015 IEEE International Conference on Computer Vision (ICCV), 2015, pp. 1395-1403.

* * *